\documentclass[letterpaper, 10 pt, conference]{ieeeconf}
\usepackage{amsmath,lipsum}
\usepackage{lipsum}
\usepackage{mathtools, nccmath}
\usepackage{cuted}
\usepackage{color}
\usepackage[keeplastbox]{flushend}
\usepackage{mathtools,amssymb,lipsum, nccmath}

\usepackage{cuted}
\usepackage{graphicx}
\usepackage{comment}
\usepackage{algorithm}
\usepackage{algorithmicx}
\usepackage{algpseudocode}
\usepackage{xcolor}
\usepackage{caption}
\usepackage{bm}
\usepackage{float}
\usepackage{url}
\usepackage{subfig}

\usepackage{url}

\IEEEoverridecommandlockouts                              
\title{\LARGE \bf Task Space Planning with Complementarity  Constraint-based Obstacle Avoidance}
\author{Anirban Sinha$^{1}$, Anik Sarker$^{1}$ and Nilanjan Chakraborty$^{1}$
}

\begin{document}
\maketitle
\thispagestyle{empty}
\pagestyle{empty}
\begin{abstract}
\noindent
{\em In this paper, we present a task space-based local motion planner that incorporates collision avoidance and constraints on end-effector motion during the execution of a task.  Our key technical contribution is the development of a novel kinematic state evolution model of the robot where the collision avoidance is encoded as a complementarity constraint. We show that the kinematic state evolution with collision avoidance can be represented as a Linear Complementarity Problem (LCP). Using the LCP model along with Screw Linear Interpolation (ScLERP) in $SE(3)$, we show that it may be possible to compute a path between two given task space poses by directly moving from the start to the goal pose, even if there are potential collisions with obstacles. Scalability of the planner is demonstrated with experiments using a physical robot. We present simulation and experimental results with both collision avoidance and task constraints to show the efficacy of our approach.}

\noindent
{\bf Keywords: Motion Planning, Task-Space Planning, Complementarity Constraints, Collision Avoidance.}

\end{abstract}

\section{Introduction}

\begin{figure}[!t]
 \centering
 \includegraphics[width=0.5\textwidth]{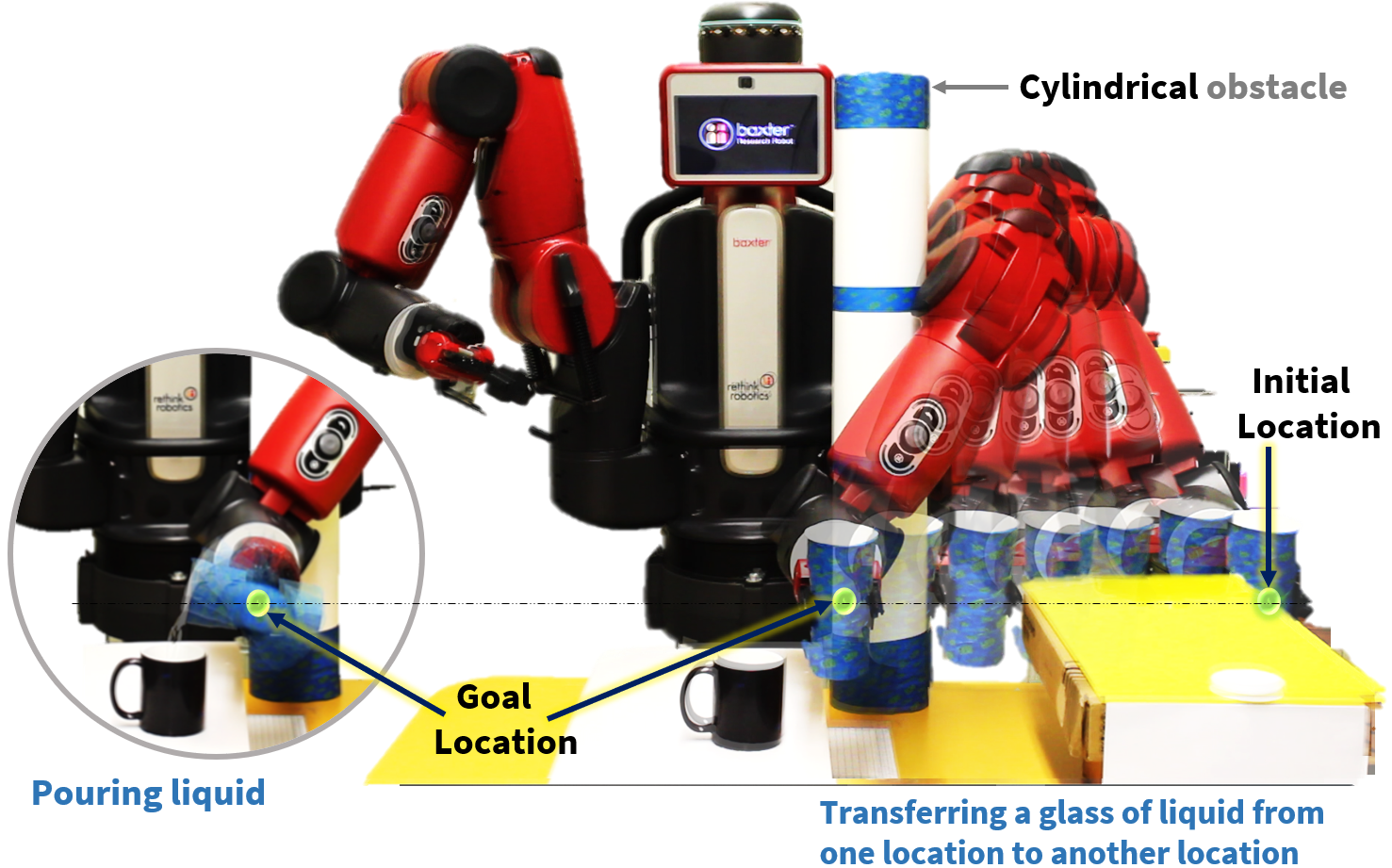}
 \caption{Path with constrained end-effector motion and simultaneous obstacle avoidance, computed using the proposed planner.}
 \label{transfer_pour_exp_front}
 \end{figure}

Planning the motion of robot manipulators to move its end-effector from an initial to a goal pose (position and orientation) while avoiding obstacles and satisfying constraints on the motion of the end effector (if any), is a fundamental problem in robotics. For example, consider the manipulation task shown in Figure~\ref{transfer_pour_exp_front}. The robot has to transfer and pour a glass of water to another container while avoiding collision with the cylindrical obstacle. To execute the task successfully, the robot has to compute a collision-free path such that the orientation of the glass needs to be held constant during the transfer to the cup location and the position of the glass should be constant during pouring of the water. These constraints on the pose of the glass (or the end effector of the robot). Therefore our goal is to develop algorithms to compute collision free motion plans where there may be constraints on the end effector of the robot.

There are multiple spaces associated with a robot and the different constraints occur in these different spaces. The robot operates in the $3D$ world or $\mathbb{R}^3$ and the collision avoidance constraints can be formulated in $\mathbb{R}^3$. The set of all poses of the robot end effector is called the {\em task space} ($\mathbb{T}$-space)  and the constraints on end effector motion as shown in the example in Figure~\ref{transfer_pour_exp_front} occurs in the task space. The task space of a robot is a subset of $SE(3)$, the group of rigid body motions. The {\em joint space} or configuration space ($\mathbb{C}$-space) of a manipulator is the set of all joint angles of the manipulator. The robot motion is controlled by controlling the joint angles and the motion plan has to be in the joint space. 

A large body of work in motion planning has been in developing configuration space-based approaches~\cite{Latombe,Lavalle,choset2005principles}. Among them sampling-based algorithms have been the most successful ones~\cite{KavrakiSLO96,LavalleK01}. These algorithms search for collision free paths in the joint space of the robot and use a collision detection algorithm (which are implemented in $\mathbb{R}^3$ using the geometry of the links of the robot and the obstacles).  However, constraints on the motion of the end effector becomes nonlinear manifolds in the joint space (because the forward kinematics maps are nonlinear). Generating samples from a manifold in a high dimensional space is usually hard and this makes incorporating motion constraints in joint space-base planning schemes hard, although there have been effort made in this direction~\cite{berenson2009manipulation,BerensonSK11,JailletP12,Stilman10,brock2000elastic,KimU16,YaoK07, bonilla2015sample, KingstonMK2019}. 

On the other hand, task-space based planning approaches~\cite{whitney1969resolved, Khatib87} usually use a linear interpolation of the parameterization used for representing $SE(3)$. This again implies that the motion constraints have to be encoded explicitly and sometimes they may give rise to nonlinear equations at the position level. Furthermore, it is harder to incorporate collision avoidance in these techniques, although potential field methods and its variants has been used for collision avoidance~\cite{Khatib86,yang2010elastic}. The planner can get stuck in a local minimum and may not get a feasible solution even if one exists.
In prior work~\cite{sarker2020}, we have shown that a large class of motion constraints in $SE(3)$, namely, those that restrict the motion to a subgroup of $SE(3)$ (or a sequence of subgroups of $SE(3)$) can be easily satisfied using Screw Linear Interpolation (ScLERP) as a task-space based planner (instead of just using linear interpolation of the parameters). However, in~\cite{sarker2020}, we did not consider collision avoidance constraints. 

The goal of the paper is to develop a $\mathbb{T}$-space based local planner that can provide collision-free paths. For such planner, search space would be of constant dimension (less than or equal to $6$) irrespective of DoF of robot and can potentially enjoy advantages of both task-space based approaches and the sampling framework of configuration space-based approaches. The local planner will leverage our work in~\cite{sarker2020} that uses Screw Linear Interpolation (ScLERP) in $SE(3)$ along with RMRC.  

Our local planner is based on two key ideas: (a) a novel kinematic state update model that relates the task space velocities to joint space velocities while also considering the collision avoidance constraint by using a complementarity-based formulation. (b) the use of Screw Linear Interpolation (ScLERP) to generate a path between two given task space configurations\cite{sarker2020}. 

Complementarity constraints have been used in the context of joint-space based motion planning with dynamics~\cite{chakraborty2009complementarity}. However, in this paper we show that complementarity constraints can also be used with kinematic state update equations for $\mathbb{T}$-space based planning. Intuitively, {\em the complementarity constraints enables us to develop a collision-free path that is guided by the obstacles boundaries} (please see the results in Section~\ref{sec:res}).

The key contributions of this paper are as follows: 
\begin{enumerate}
    \item We develop a novel kinematic motion model that combines the differential kinematics of a manipulator along with a complementarity-based formulation for avoiding obstacles. The discrete-time motion model is a linear complementarity problem (LCP) that can be used as a local planner in $SE(3)$ that always generates collision free paths.
    \item We show that although our algorithm is a local algorithm it can generate feasible paths to the goal in a cluttered environment (that may not be star-shaped) by moving directly towards the goal.
    \item We also show empirically through experiments that the use of ScLERP ensures that certain classes of task space constraints are satisfied without explicitly enforcing them. The theoretical discussion on why ScLERP can encode some common constraints on $SE(3)$ is presented in our earlier work~\cite{sarker2020}, so we will not discuss it in any detail here.
\end{enumerate}

\section{Related Work} \label{sec:rw}
The extant literature on motion planning in robotics is quite extensive~\cite{Latombe,Lavalle,choset2005principles}. Since, we are proposing a planner in $\mathbb{T}$-space for collision avoidance,  we will focus on task space based planning and sampling-based techniques that are geared towards solving narrow passage and constrained motion problems.

The extant literature on motion planning in $\mathbb{T}$-space with task constraints and collision avoidance constraints can be divided into two broad categories. The first type is based on describing a continuous path in $\mathbb{T}$-space~\cite{ambler1975inferring} satisfying task constraints and computing corresponding joint space path using the method of resolved rate motion control~\cite{whitney1969resolved}\cite{AnthonyA}. A related work~\cite{Mao2019} considers real-time dynamic planing where task constraints are optimized and the planner changes its goal location on the fly. However, in a more cluttered environment this changing goal on the fly may incur in-feasibility. A planner ensuring desired end-effector force and pose constraints is proposed in~\cite{Khatib87}. In order to simultaneously maintain end-effector task constraint and obstacle avoidance, null-space motion of the end-effector is utilized. For obstacle avoidance with this approach, the proposed potential field based method~\cite{Khatib86} drives the robot along the gradients of the predefined potential functions that ensures avoiding the obstacles. However this approach suffers from getting stuck in local minimum of the potential fields and there is no general method to design such potential fields that has only one global minimum. On the other hand our approach do not need to design obstacles with potential functions and no pre-computed task space path is required. Our planner computes end-effector path iteratively using ScLERP based interpolation. Although the proposed planner can also get stuck, but it can find plans for cases where potential functions or navigation functions~\cite{RimonK92,yang2010elastic} based planners cannot (see example $1$ in section~\ref{sec:res}). Furthermore, it is quite easy to incorporate our planner within a sampling-based planning framework. Penalty function based approach to compute optimal trajectories of redundant robots is proposed in~\cite{zucker2013chomp,schulman2014motion,dong2016motion,kalakrishnan2011stomp}. The penalty functions are composed with relaxed constraints obtained by softening the hard task-space geometric constraints. However this approach does not perform efficiently if the number of constraints or the degrees of freedom of the robot is high.

\textbf{Motion Planning in} $\mathbb{C}$-\textbf{space with task constraints}:
Motion planning in $\mathbb{C}$-space is mostly done by sampling based techniques. The variety of sampling based approaches can be divided into a few subsets as follows: (a) relaxation of the constraint functions allowing a non-zero tolerance of a sample's distance from the constraint manifold. This approach is equivalent to generating configurations in narrow passage\cite{bialkowski2013free}. Once a close-to-satisfying configurations is found, a standard local planner can be utilized to generate edge of the tree being grown as presented in~\cite{bonilla2015sample, bonilla2017noninteracting,rodriguez2008resampl}. However path obtained from this planning method does not ensure execution of the path as it grossly dependent on the capability of controller being used for trajectory execution. (b) Projecting a randomly generated configuration to the constraint manifold that satisfies task constraint is the most applied method in sampling based planners with constraints\cite{yao2005path,Stilman10}. The general idea is to iteratively projecting an end-effector configuration to the constraint manifold using the gradient (commonly pseudo-inverse of the Jacobian of the constraints) of the constraint function until a joint configuration is obtained that satisfies all the constraints up to a given error tolerance.~\cite{berenson2009manipulation} proved probabilistic completeness and introduced a bidirectional tree version of this approach. (c) Planning in the tangent-space of constraint manifold by locally approximating the constraint manifold are done in~\cite{Stilman10,weghe2007randomized}. Bases of tangent space are defined by the vectors in the null space of the constraint Jacobian which require decomposition of the Jacobian matrix with the expense of computational burden. \cite{mcmahon2018sampling} introduced Reachable Volume (RV) for sampling based motion planning with task constraints. They defined RV as a space, in which any point will automatically satisfy the task constraints. They show that planning a constrained motion in RV space is similar to planning for unconstrained motion. 

\section{Mathematical Preliminaries}
\label{sec:prelim}
\textbf{Screw Linear Interpolation (ScLERP):}
Let ${\bf A} = {\bf A}_{\rm r} + \frac{\epsilon}{2}{\bf t}_{\rm A}\otimes{\bf A}_{\rm r}$ and ${\bf B} = {\bf B}_{\rm r} + \frac{\epsilon}{2}{\bf t}_{\rm B}\otimes{\bf B}_{\rm r}$ are two unit dual quaternions corresponding to two rigid body configurations between which we need to generate interpolated poses. Then the transformation ${\bf A}^{*}\otimes{\bf B}$ implicitly describes the distance in position and orientation about the screw axis. Let this distance in position and orientation be denoted as $\rm d$ and $\rm \theta$ respectively. Using ScLERP, we can generate intermediate poses between ${\bf A}$ and ${\bf B}$ based on screw motion. 
Also the parameter $\rm \tau$ weights $\rm d$ and $\rm \theta$ each with a weight of $\rm \tau$ to generate the new interpolated pose ${\bf C}(\rm \tau)$. The general expression in compact notation to compute ${\bf C}(\rm \tau)$ is
\begin{equation} \nonumber
    {\bf C}(\rm \tau) = {\bf A}\otimes({\bf A}^*\otimes{\bf B})^{\rm \tau} \quad \rm \tau \in [0, 1]
\end{equation}
where $\otimes$ denote dual quaternion multiplication (please see~\cite{shoemake1985,daniilidis1999hand} for details).

\textbf{Complementarity problem:}
\label{sec:intro_complementarity}
In the paper we will present {\em continuous collision free state evolution model} which forms a {\em differential complementarity problem}(DCP)~\cite{anitescu1997formulating,facchinei2007finite,tzitzouris2002numerical} to model the contact constraint. Let ${\bf u} \in \mathbb{R}^{\rm n_1}$, ${\bf v} \in \mathbb{R}^{\rm n_2}$. Also let two vector valued functions ${\bf g}:\mathbb{R}^{\rm n_1}\times\mathbb{R}^{\rm n_2}\rightarrow\mathbb{R}^{\rm n_1}$, ${\bf f}:\mathbb{R}^{\rm n_1}\times\mathbb{R}^{\rm n_2}\rightarrow\mathbb{R}^{\rm n_2}$. Let the notation $0\leq {\bf x} \perp {\bf y} \geq 0$ imply, ${\bf x}$ is orthogonal to ${\bf y}$ and each component of the vectors is non-negative. 

\textbf{Definition 1:} The DCP is to find ${\bf u}, {\bf v}$, satisfying
$\dot{\bf u}={\bf g}({\bf u},{\bf v})$ and $0\leq{\bf v}\perp{\bf f}({\bf u},{\bf v})\geq0$.

\textbf{Definition 2:} The MCP is to find ${\bf u}, {\bf v}$, satisfying
${\bf g}({\bf u},{\bf v})=0$ and $0\leq{\bf v}\perp{\bf f}({\bf u},{\bf v})\geq0$ .
When ${\bf f}$ and ${\bf g}$ are linear, the problem is called {\em mixed linear complementarity problem} (MLCP), otherwise, {\em mixed nonlinear complementarity problem} (MNCP).

\section{Problem Statement}
\label{sec:ps}
Let ${\bf g}(0) = {\bf g}_{0} \in SE(3)$ be the initial and ${\bf g}_{d} \in SE(3)$ be the desired end-effector poses. Let $\bm{\Theta}(0) \in \mathbb{R}^{n}$ be the initial joint configuration vector of a $n$-joint manipulator, and $\theta_{k}$ be the $k^{\rm th}$ joint angle. We want to compute a path in joint space as a sequence of joint angle vectors, $\bm{\Theta}(i)$, $i =  0, \dots, m$, such that $\mathcal{FK} (\Theta ( m)) = {\bf g}_{d}$, where $\mathcal{FK}$ is the forward kinematics map. The path should also satisfy all or a subset of the following three types of constraints: 

(a) {\em Joint Limits}: Each joint angle in the plan should satisfy $\theta_{k}^{\rm min} \leq \rm \theta_{k} (i) \leq \theta_{k}^{\rm max}$, for all $i, k$, where $\theta_{k}^{\rm min}$ and $\theta_{k}^{\rm max}$ are the lower and upper bounds of the $k^{th}$ joint.

(b) {\em Collision Avoidance}: For each $\bm{\Theta}(i)$, none of the manipulator links should collide with the obstacles. The collision avoidance constraints are naturally expressed and computed in the Euclidean world $\mathbb{R}^3$. Expressing the collision avoidance constraints in the $\mathbb{T}$-space is ill-posed since for the same end-effector pose there may be achieved with multiple inverse kinematics solutions, some of which may avoid obstacles for all the links but some may not. Expressing the collision avoidance constraints in the joint space is mathematically well-posed, however, it is computationally hard to have a joint space representation of the obstacles. 

(c) {\em Task Constraints}: These are constraints on the end-effector motion specific to the task. They can be naturally expressed in the $\mathbb{T}$-space. Task constraints can also be expressed in joint space by constraining the motion of each joint movements resulting in many constraint equations proportional to the number of joints.

\section{State Evolution with Collision Avoidance}
\label{sec:kin_mod}
The proposed motion model that computes path and avoid obstacles simultaneously is composed of two fundamental components (a) a ScLERP based kinematic state evoultion model that was introduced in out previous work~\cite{sarker2020,sinha2020task} and (b) complementarity constraint to model collision constraints. 

\textbf{Kinematic Motion Model}:
Let ${\bf p}$ denote position of the end effector and ${\bf Q}$ be the unit-quaternion representing orientation of the end effector. Let ${\bf v}$, ${\bf \omega}$ are the spatial linear and angular velocities of the end-effector respectively. Let ${\bf V} = [{\bf v}^{\rm T}, {\bf \omega}^{\rm T}]^{\rm T}$. Then velocity kinematics relationship is
\begin{equation}
\label{eq: thetadot2V}
   {\bf V} = {\bf J} \dot{\bm{\Theta}} 
\end{equation}
where $\dot{\bm{\Theta}}$ is the vector of joint velocities and ${\bf J}$ is the   manipulator Jacobian. Let ${\bf \gamma} = [{\bf p}^{\rm T} \quad {\bf Q}^{\rm T} ]^{\rm T}$ and $\dot{{\bf \gamma}} = [\dot{\bf p}^{\rm T} \quad \dot{\bf Q}^{\rm T}]^{\rm T}$ The velocity ${\bf V}$ is related to $\dot{\bf p}$ and $\dot{\bf Q}$ as
\begin{equation}
\label{eq:repJac}
    {\bf V} = {\bf J}_{\rm r}\dot{{\bf \gamma}} 
\end{equation}
where ${\bf J}_{\rm r}$ is the $6\times7$ representation Jacobian. Note that the manipulator Jacobian in Equation~\eqref{eq: thetadot2V} can be spatial, body or analytic Jacobian~\cite{murray2017mathematical}. The representation Jacobian expression will depend on the choice of manipulator Jacobian. 
From Equations~\eqref{eq: thetadot2V} and~\eqref{eq:repJac} we get the relationship between $\mathbb{T}$-space velocities and joint rates as
\begin{equation}
    \label{eq:vkin1}
    \dot{\bm{\Theta}} = {\bf J}^{\rm T} \left( {\bf J} {\bf J}^{\rm T}\right)^{-1}
    {\bf V} \\
    = {\bf J}^{\rm T} \left( {\bf J} {\bf J}^{\rm T}\right)^{-1}{\bf J}_{\rm r}
    \dot{\bm{\gamma}} = {\bf B}\dot{{\bf \gamma}}
\end{equation}
where ${\bf B} = {\bf J}^{\rm T} \left({\bf J} {\bf J}^{\rm T}\right)^{-1}{\bf J}_{\rm r}$. 
For a $\rm n$-DoF manipulator,  in Equation~\eqref{eq:vkin1}, $\dot{\bm{\Theta}}\in\mathbb{R}^{\rm n}$, ${\bf B}\in\mathbb{R}^{\rm n\times7}$ and $\dot{{\bf \gamma}}\in\mathbb{R}^{7\times1}$. Equation~\eqref{eq:vkin1} can be used for both kinematics based motion planning and inverse kinematics (or redundancy resolution)  for redundant manipulators. Let $\bm{\Theta}^{\rm t}$ and ${\bf \gamma}^{\rm t}$ represent joint angle vector and corresponding end-effector pose respectively.

\textbf{Motion Model for Obstacle Avoidance}: 
We will now modify the motion model in Equation~\eqref{eq: thetadot2V} to model obstacle avoidance in a cluttered environment. To do that we will first assume that we have some method to obtain a compensating Cartesian space velocity of magnitude $v_{\rm c_{\rm i}}$ along the contact normal $\bm{N}_{c_i}$ whenever any link of the manipulator (say the $i^{th}$ link) comes in virtual contact with any of the obstacles in the environment. Let $\bm{J}_{\rm c_{\rm i}}$ be the Jacobian upto the contact point on link $i$. Then the additional joint rate imposed by the compensating velocity at the virtual contact can be obtained as
\begin{equation}
    \label{eq:theta_add}
    \dot{\bm{\Theta}}_{\rm add} = {\bf J}_{\rm c_{\rm i}}^\dagger {\bf N}_{\rm c_{\rm i}} \rm v_{\rm c_{\rm i}}
\end{equation}

Thus the required joint rate that brings the end-effector towards the desired goal and also avoids obstacles in the environment can be obtained by combining the joint rates from Equation~\eqref{eq:vkin1} and~\eqref{eq:theta_add} together as presented in Equation~\eqref{eq:th_combined}
\begin{equation}
    \label{eq:th_combined}
    \dot{\bm{\Theta}} = {\bf B}\dot{{\bf \gamma}} + {\bf J}_{\rm c_{\rm i}}^\dagger {\bf N}_{\rm c_{\rm i}} \rm v_{\rm c_{\rm i}}
\end{equation}

We can extend Equation~\eqref{eq:th_combined} for the case when multiple links are in virtual contacts with the obstacles, by simply adding the effects of all the compesating Cartesian velocities on the joint rates as in Equation~\eqref{eq:req_joint_rate}.
\begin{equation}
    \label{eq:req_joint_rate}
    \dot{\bm{\Theta}} = {\bf B}\dot{{\bf \gamma}} + \sum_{i=1}^{\rm n_{c}} {\bf J}_{\rm c_{\rm i}}^\dagger {\bf N}_{\rm c_{\rm i}} \rm v_{\rm c_{\rm i}}
\end{equation}
where $n_c$ is the total number of links in virtual contact with the obstacles.Note that in Equation~\eqref{eq:req_joint_rate}, we are projecting each compensating velocity at a contact on link $i$ to the entire joint space of the manipulator up to the joint constraining link $i$ and $i-1$. Thus, the actual task space velocity would be different from the desired input task space velocity $\dot{{\bf \gamma}}$. 
\vskip 0.1 in

However, it is possible to achieve the same input task space velocity at the end effector by projecting the compensating velocity $(\sum_{i=1}^{\rm n_{c}} {\bf J}_{\rm c_{\rm i}}^\dagger {\bf N}_{\rm c_{\rm i}} \rm v_{\rm c_{\rm i}})$ into the null space of the manipulator Jacobean. 
There can be infinitely many joint velocity vectors which will not affect the end-effector velocity. This would imply that the end-effector velocity would be the same as the desired input task space velocity $\dot{{\bf \gamma}}$. With the addition of null
space joint velocities, the relationship between end-effector velocity and manipulator joint velocities takes the following form  
\begin{equation}
    \label{eq:null_continuous_model}
    \dot{\bm{\Theta}}  =  {\bf B} \dot{\bf \gamma} +(\bf I-  {\bf J}_{\rm e}{\bf J}_{\rm e}^\dagger) \sum_{i=1}^{n_{\rm c}}{\bf J}_{\rm c_{\rm i}}^\dagger {\bf N}_{\rm c_{\rm i}} \rm v_{\rm c_{\rm i}}
\end{equation}
Here, ${\bf J}_{\rm e}$ is the manipulator Jacobean and ${\bf J}_{\rm e}^\dagger$ is a generalized inverse of ${\bf J}_{\rm e}$.  ${\bf J}_{\rm e}^\dagger$. 

\textbf{Complementarity Constraints for Collision Modeling}:
\label{sec:contact_modeling}

Complementarity constraints have been used to model non-penetration contact constraints in rigid body dynamics~\cite{chakraborty2009complementarity}. However in this paper we show that complementarity constraint can also be used to model contact at the kinematic level in terms of compensating velocities to avoid obstacles. Next we will give a brief description on how complementarity constraint can be used to model kinematic collision constraints with the help of examples with a point and a $2R$ robot and then we will show how the same idea can be utilized for higher dimension cases. 

\begin{figure}[!htbp]
 \centering
    \includegraphics[width=0.41\textwidth]{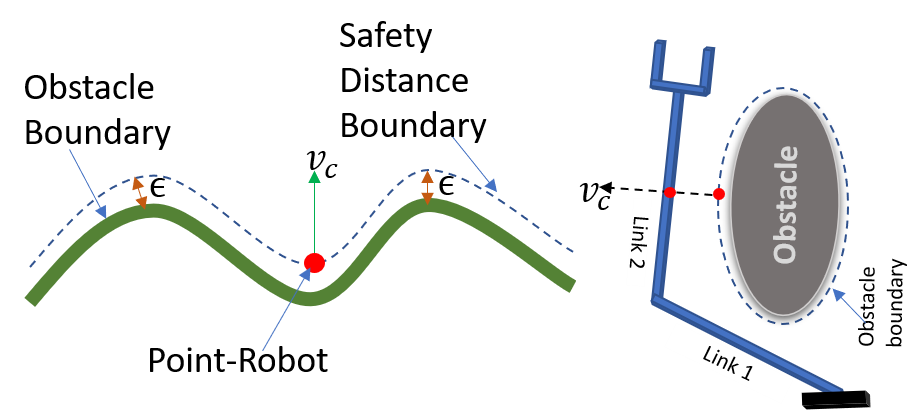}
 \caption{A point robot (left) and a $2$R robot link (right) in the vicinity of the $\epsilon$ distance from the obstacle, experience compensating velocity $\rm v_{\rm c}$ that ensure avoiding the obstacles.}
\label{fig:concept_conpensating_vel}
\end{figure}

Consider a point robot approaching an obstacle (see schematic sketch in Figure~\ref{fig:concept_conpensating_vel}), Let ${\bf q}^{\rm t}$ denotes configuration of the robot and $\rm \psi({\bf q}^{\rm t})$ be a signed distance function between the point robot and the obstacle as a function of the robots configuration. When the distance between the robot and the obstacle is $>\rm 0$ (or $> \epsilon$ if a buffer around the obstacles is considered), there is no collision, and hence the compensating velocity $v_c = \rm 0$. In contrary, when the gap distance is $\rm 0$, a non-zero compensating velocity comes into play ensuring no collision. \footnote{When both $\rm v_{\rm c} = 0$ and $\rm \psi({\bf q}^{\rm t}) = 0$, it implies that the point is in grazing contact with the object with no normal velocity component.} Mathematically, this constraint can be written as a complementarity constraint as $0 \leq \rm v_{\rm c} \perp \rm \psi({\bf q}^{\rm t}) \geq 0$.
Similarly for a the $2R$ manipulator whose second link is in virtual contact with the obstacle (please see right of Figure~\ref{fig: point_robot1}), the complementarity constraint between the compensating velocity and the arm configuration can be written as $0 \leq \rm v_{\rm c} \perp \rm \psi(\bm{\Theta}^{\rm t}) \geq 0$ where $\Theta^t$ is the current joint configuration of the robot.

\textbf{Kinematic State Evolution Model with Complementarity Constraint-based Obstacle Avoidance}:
Combining kinematic state evolution model and complementarity constraint for obstacle avoidance, we get kinematic state evolution model ensuring collision avoidance as,
\begin{eqnarray} \nonumber
    \dot{\bm{\Theta}} & = & {\bf B}\dot{{\bf \gamma}} + \sum_{\rm i=1}^{\rm n_c}{\bf J}_{\rm c_{\rm i}}^\dagger {\bf N}_{\rm c_{\rm i}} \rm v_{\rm c_{\rm i}}\\
    0  & \leq &  \rm v_{\rm c_{\rm i}} \perp {\rm \psi}_{\rm i}(\bm{\Theta}) - \epsilon \geq 0
    \label{eq:continuous_model}
\end{eqnarray}
The model above is a Differential Complementarity Problem (DCP) where the unknowns are the joint angle vector $\bm{\Theta}$ and the compensating velocities $\rm v_{\rm c_{\rm i}}$'s. Any solution of Equation~\eqref{eq:continuous_model} will ensure that all the robot links are at least $\epsilon$ distance away from any of the obstacles present in the workspace. A discrete-time version of this state evolution model, which we present below, forms the basis of our local planner as presented in section~\ref{sec:planning_algorithm}.  

We will use a backward Euler time-stepping scheme to obtain the discrete time equations, which is a Nonlinear Complementarity Problem (NCP) as in Equation~\eqref{eq:discrete_model}, is one of the key contribution of the paper.
\begin{eqnarray} \nonumber
    \bm{\Theta}^{\rm t+h} & = & \bm{\Theta}^{\rm t} + {\bf B}({\bf \gamma}^{\rm t+h} - {\bf \gamma}^{\rm t})+ \rm h \sum_{\rm i=1}^{\rm n_{\rm c}}{\bf J}_{\rm c_{\rm i}}^\dagger {\bf N}_{\rm c_{\rm i}} \rm v_{\rm c_{\rm i}}^{\rm t+h}\\
    \label{eq:discrete_model}
    0 & \leq & \rm v_{\rm c_{\rm i}}^{\rm t+h} \perp \rm \psi_{\rm i}(\bm{\Theta}^{\rm t+h}) - \epsilon \geq 0
\end{eqnarray}
where $h$ is the discretization step length. We can further linearize the distance function $\rm \psi_{\rm i}^{\rm t+h} \approx \rm \psi_{\rm i}^{\rm t}+h\frac{\partial \rm \psi}{\partial \rm t}=\rm \psi_{\rm i}^{\rm t} + \rm h {\bf N}_{\rm c_{\rm i}}^{\rm T}{\bf J}_{\rm c_{\rm i}}(\bm{\Theta}_{\rm r}^{\rm t+h} - \bm{\Theta}_{\rm r}^{\rm t})$ (first order Taylor's series). Then discrete time state evolution model with complementarity based obstacle avoidance to convert the NCP in Equation~\eqref{eq:discrete_model} in to a Linear Complementarity Problem (LCP) as presented in Equation~\eqref{eq:discrete_model1}. Since LCP can be solved faster than NCP, this will be advantageous to get solutions faster. 
\begin{eqnarray} \nonumber
    \label{eq:discrete_model1}
    \bm{\Theta}^{\rm t+h} = \bm{\Theta}^{\rm t} + {\bf B}({\bf \gamma}^{\rm t+h} - {\bf \gamma}^{\rm t})+ \rm h \sum_{\rm i=1}^{\rm n_{\rm c}}{\bf J}_{\rm c_{\rm i}}^\dagger {\bf N}_{\rm c_{\rm i}} \rm v_{\rm c_{\rm i}}^{\rm t+h}\\
    0 \leq \rm v_{\rm c_{\rm i}}^{\rm t+h} \perp \rm \psi_{\rm i}^{\rm t} + \rm h {\bf N}_{\rm c_{\rm i}}^{\rm T}{\bf J}_{\rm c_{\rm i}}(\bm{\Theta}^{\rm t+h} - \bm{\Theta}^{\rm t}) - \epsilon \geq 0
\end{eqnarray}

\section{Uniqueness Property of the Solution}
Since our motion model is a LCP, solving which is a NP-hard problem in general~\cite{cottle1992linear}, a question arises whether the LCPs that are generated here are NP-hard. Here we prove that if a link has single point contact or two-point contact then the LCP in Equation~\eqref{eq:discrete_model1} has unique solution that can be computed in polynomial time. Writing the two equations in~\eqref{eq:discrete_model1} into one by eliminating $\bm{\Theta}^{\rm t+h}$ and $\bm{\Theta}^{\rm t}$ we get Equation~\eqref{eq:only_vc} with only unknowns being $\rm v_{\rm c_i}$s.
\begin{equation}
\label{eq:only_vc}
0 \leq \rm v_{\rm c_i}^{\rm t+h} \perp \psi_{\rm i}^{\rm t} + \rm h {\bf N}_{\rm c_i}^{\rm T}{\bf J}_{\rm c_i}{\bf K}
\end{equation}
where ${\bf K} = {\bf B}({\bf \gamma}^{\rm t+h} - {\bf \gamma}^{\rm t}) + h\sum_{i=1}^{\rm n_c}{\bf J}_{\rm c_i}^\dagger{\bf N}_{\rm c_i}v_{\rm c_i}^{\rm t+h}$. Notice that the Equation~\eqref{eq:only_vc} has the form $ {\bf 0} \leq {\bf z} \perp {\bf q} + {\bf Mz} \geq {\bf 0}$,
where, ${\bf z},{\bf q} \in \mathbb{R}^{n_c}$ and the corresponding $i^{th}$ elements are $\rm v_{c_i}$'s and $\psi_{\rm i}^{\rm t} + \rm h {\bf N}_{\rm c_i}^{\rm T}{\bf J}_{\rm c_i}{\bf B}({\bf \gamma}^{\rm t+h} - {\bf \gamma}^{\rm t})$ respectively, where $i \in 1\dots n_c$. The matrix ${\bf M} \in \mathbb{R}^{n_c \times n_c}$ where $m_{ij}=\rm h^2 {\bf N}_{\rm c_i}^{\rm T}{\bf J}_{\rm c_i}{\bf J}_{\rm c_j}^\dagger{\bf N}_{\rm c_j}$ where $i,j \in 1\dots n_c$.
Following~\cite[pp.141]{cottle1992linear}, if the matrix $\bf M$ is {\em positive definite}, then a LCP has unique solution.
\subsubsection*{One Contact Case: } Here $n_c=1$, then matrix $\bf M$ contains only one element $\rm m_{11}=\rm h^2{\bf N}_{\rm c_1}^{\rm T}{\bf J}_{\rm c_1}{\bf J}_{\rm c_1}^\dagger{\bf N}_{\rm c_1} = \rm h^2 > 0$. That means, for one contact point, the matrix $\bf M$ is always {positive definite}, hence unique solution exists.
\subsubsection*{Two Contact Case: } If one link of the manipulator experiences two contacts, then also we can show the corresponding LCP has unique solution. In this case the matrix $\bf M$ will be of dimension $2 \times 2$ and we need to show that $\bf M$ is {\em positive definite} for the uniqueness of the solution. We can show,
\begin{equation}
    \text{det}|{\bf M}| = {\rm m}_{11}{\rm m}_{22} - {\rm m}_{21}{\rm m}_{12} = {\rm h}^4\left(1 - \left(\hat{\bf n}_{\rm c_1}^{\rm T}\hat{\bf n}_{\rm c_2}\right)^2\right)
\end{equation}
where $\hat{\bf n}_{\rm c_i}\in \mathbb{R}^3$ is the unit normal vector at the $i^{\rm th}$ contact point. Therefore as long as $\hat{\bf n}_{\rm c_1}$ and $\hat{\bf n}_{\rm c_2}$ are not parallel, $\hat{\bf n}_{\rm c_1}^{\rm T}\hat{\bf n}_{\rm c_2}<1$. Hence $\text{det}|{\bf M}|>0$ or ${\bf M}$ is {\em positive definite}. 

Although the results presented above are not completely general, this is sufficient for handling virtual contacts because we can handle multiple simultaneous contacts sequentially. 
\begin{algorithm}[!htb]
\caption{LCP based Local Planner\\ 
\textbf{Input}: ${\bf \gamma}_{\rm start}$, ${\bf \gamma}_{\rm goal}$, $\bm{\Theta}_{\rm start}$, $\rm \tau$  \, \textbf{Output}: ${\bf \gamma}_{\rm new}$, $\bm{\Theta}_{\rm arr}$}
\label{alg3}
\begin{algorithmic}[1]
\State ${\bf \gamma}_{\rm old} \leftarrow {\bf \gamma}_{\rm start}$, $\bm{\Theta}_{\rm old} \leftarrow \bm{\Theta}_{\rm start}$, $\bm{\Theta}_{\rm arr} \leftarrow [\,]$
\State ${\bf dq}_{\rm goal} \leftarrow \text{VEC2DQ}({\bf \gamma}_{\rm goal})$
\While {not reached}
    \State ${\bf dq}_{\rm old} \leftarrow \text{VEC2DQ}({\bf \gamma}_{\rm old})$
    \State ${\bf dq}_{\rm new} \leftarrow \text{ScLERP}({\bf dq}_{\rm old}, {\bf dq}_{\rm goal}, \rm \tau)$
    \State ${\bf \gamma}_{\rm new} \leftarrow \text{DQ2VEC}({\bf dq}_{\rm new})$
    \State Solve Eq.\eqref{eq:discrete_model1} to get joint angle vector $\bm{\Theta}_{\rm new}$
    \State Update ${\bf \gamma}_{\rm new}$ as ${\bf \gamma}_{\rm new} \leftarrow \mathcal{FK}(\bm{\Theta}_{\rm new})$
    \If {DIST$({\bf \gamma}_{\rm new}, {\bf \gamma}_{\rm goal}) < \text{tol}$ \text{OR} DIST(${\bf \gamma}_{\rm new}, {\bf \gamma}_{\rm old}) < \text{tol}$}
        \State \Return ${\bf \gamma}_{\rm new}, \bm{\Theta}_{\rm arr}$
    \EndIf
    \State ${\bf \gamma}_{\rm old} \leftarrow {\bf \gamma}_{\rm new}$, $\bm{\Theta}_{\rm old} \leftarrow \bm{\Theta}_{\rm new}$
    \State $\bm{\Theta}_{\rm arr}.{\rm Append}(\bm{\Theta}_{\rm new})$
\EndWhile
\end{algorithmic}
\end{algorithm}

\section{Summary of the proposed local planner} \label{sec:planning_algorithm}

Algorithm~\ref{alg3} gives the pseudocode of our proposed local planner. We use ${\bf dq}$ to represent a pose as a unit dual quaternion and $\bm{\gamma}$ to represent a pose as a $7 \times 1$ concatenated vector of position and orientation (represented as unit quaternion).
The input to Algorithm~\ref{alg3} is the initial pose, ${\bf \gamma}_{\rm start}$, goal pose, ${\bf \gamma}_{\rm goal}$, initial joint vector $\bm{\Theta}_{\rm start}$ (inverse kinematics solution~\cite{sinha2019geometric} for ${\bf \gamma}_{\rm start}$) and interpolation parameter $\rm \tau$. The algorithm returns a sequence of joint angle vectors corresponding to the feasible path. In line $1$, the two variables ${\bf \gamma}_{\rm old}$ and $\bm{\Theta}_{\rm old}$ are initialized to ${\bf \gamma}_{\rm start}$, $\bm{\Theta}_{\rm start}$ respectively while in line $2$ we convert the ${\bf \gamma}_{\rm goal}$ into dual-quaternion form using the method VEC2DQ. Then a path is computed iteratively from line $3$ to line $13$. Line $4$ converts ${\bf \gamma}_{\rm old}$ into its dual-quaternion form ${\bf dq}_{\rm old}$. A new interpolated pose, ${\bf dq}_{\rm new}$ is computed in line $5$ using ScLERP and it is transformed into a $7\times1$ vector ${\bf \gamma}_{\rm new}$ using the method DQ2VEC. Using ${\bf \gamma}_{\rm new}$, ${\bf \gamma}_{\rm old}$, $\bm{\Theta}_{\rm old}$ (${\bf \gamma}^{\rm t+h}$, ${\bf \gamma}^{\rm t}$, and $\bm{\Theta}^{\rm t}$
Equation~\eqref{eq:discrete_model1} is solved for $\bm{\Theta}_{\rm new}$ (i.e., $\bm{\Theta}^{\rm t+h}$) and compensating velocities, $\rm v_{\rm c_{\rm i}}$ in line $7$. Line $8$ updates ${\bf \gamma}_{\rm new}$ based on $\bm{\Theta}_{\rm new}$. In line $9$, we check if the goal is reached or progress is made as compared to the previous iteration using a separate metric for $\mathbb{R}^3$ and $SO(3)$. Since there is no bi-invariant metric on $SE(3)$, we check whether position and orientation distances are smaller than a predefined threshold value independently. If any of the termination criteria is met, the Algorithm returns ${\bf \gamma}_{\rm new}$, $\bm{\Theta}_{\rm arr}$ in line $10$. Otherwise ${\bf \gamma}_{\rm old}$ and $\bm{\Theta}_{\rm old}$ are overwritten with ${\bf \gamma}_{\rm new}$ and $\bm{\Theta}_{\rm new}$ respectively to be used in the next iteration.

As indicated in the description of the Algorithm, since the proposed algorithm is a local planner, it may get stuck. This happens when the compensating velocity at the contacts is equal and opposite to the velocity of the contact point on the robot. This is to be expected since it is well known that motion planning is P-SPACE hard~\cite{Latombe}. However, the local planner can be used along with a sampling-based global planner adapted to $SE(3)$. In particular we use the RRT planner~\cite{LavalleK01} as the sampling-based planner in this paper. Because of space constraints and since the RRT planner is well documented in text books~\cite{Lavalle} we do not provide the details here.

\section{Simulation and Experimental Results}
\label{sec:res}
In this section we present simulation and experimental results for exemplar planning problems. The first example is for a point robot in $2D$ space moving through a narrow maze, where no direct path exists from start to goal. The environment is non-convex and not star-shaped. Therefore, potential functions~\cite{Khatib86} and navigation function~\cite{RimonK92} based methods fail to solve this problem. However, the proposed local planner can find a path by moving directly towards the goal. The second example shows that a planar $8$DoF robot moving in a cluttered environment can find a path while moving directly towards the goal. The planning is done in $2D$ task space instead of a $8$D $\mathbb{C}$-space. The third example illustrates the benefit of using ScLERP as a interpolation scheme utilized by the proposed planner. We show for water transferring task that inserting an intermediate goal deterministically can simplify the whole planning problem into two sub-problems. 
The first sub-problem constrains the end-effector to have fixed orientation so that the cup always stays upright ensuring no spillage of water. The second sub-problem constrains the end-effector position so that no water is poured outside the mug. This example is an application of robot's nullspace motion along with task constrained motion and obstacle avoidance. Note that all the timings reported in this paper are based on non-optimal MATLAB codes executed on a intel-$i5$ processor system for the purpose of proof of concept. All the planning examples presented here can be seen in action by following the link \textcolor{red}{\url{https://tinyurl.com/cmplementaritySclerp}}.


\begin{figure}[!htbp]
    \centering
    \includegraphics[width=0.5\textwidth]{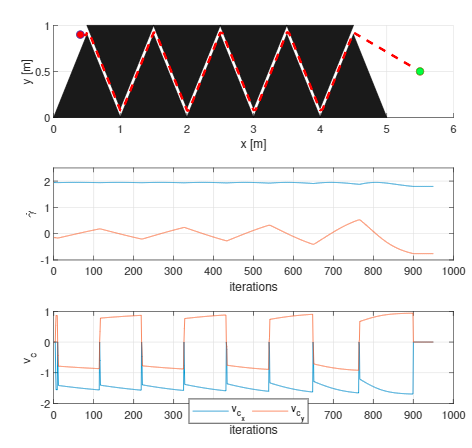}
    \caption{(Top) Computed path (red dashed line) of the point robot through the narrow maze obtained using the proposed planner. (Middle) Trajectories of the input task space velocity. (Bottom) Compensating velocity for obstacle avoidance.}
    \label{fig: point_robot1}
\end{figure}
\begin{figure}[!htbp]
    \centering
    \includegraphics[width=0.45\textwidth]{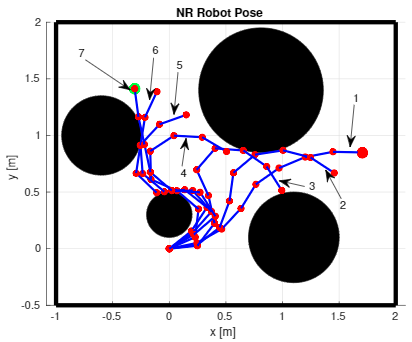}
    \includegraphics[width=0.45\textwidth]{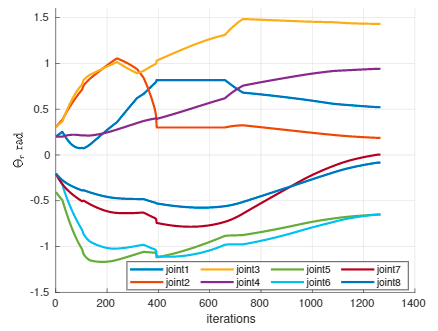}
    \caption{Top: Path of 8R-robot while moving from start (red circle) to the goal (green circle). Black circles represent obstacles. Bottom: Visualization of the plan in robot's joint space.}
    \label{fig:snake_plan}
\end{figure}
    
\textbf{Planning through narrow maze for a point robot:}\label{sec:example_point_robot}
Let the $2\rm D$ position vector of the point robot  be denoted by $[\rm x, \rm y]^{\rm T} = {\bf X} \in \mathbb{R}^2$. Noting the fact that a point robot can only make one contact with the maze wall, the collision free state evolution model in Equation~\eqref{eq:discrete_model1} becomes
\begin{eqnarray} \nonumber
    &&{\bf X}^{\rm t+h} = {\bf X}^{\rm t} + \rm h\left({\bf u}^{\rm t} + {\bf N}_{\rm c}\rm v_{\rm c}^{\rm t+h} \right)\\
    \label{eq: point_robot}
    &&0 \leq \rm v_{\rm c}^{\rm t+h} \perp \rm \psi_1({\bf X}^{\rm t}) + {\bf N}_{\rm c_1}^{\rm T} ({\bf X}^{\rm t+h} - {\bf X}^{\rm t}) \geq \epsilon
\end{eqnarray}
Note that in Equation~\eqref{eq: point_robot}, the spatial and contact Jacobian terms are not explicitly written as they will be $2\time2$ identity matrices for a point robot. Since only one possible contact can be made, Equation~\eqref{eq: point_robot} will have only one complementarity constraint equation. The input to the robot is the velocity towards the goal, ${\bf u} = [\rm v_x, \rm v_y]^{\rm T}$. Equation~\eqref{eq: point_robot} is a system of three equations with the three unknowns being ${\bf X}^{\rm t+h} \in \mathbb{R}^2$ and $\rm v_{\rm c}^{\rm t+h}\in\mathbb{R}$. The input towards the goal at step $t$ is  ${\bf u}^t = {\bf K}_{\rm p}({\bf X}_{\rm G} - {\bf X}_{\rm r}^{\rm t})/(\rm h||{\bf X}_{\rm G} - {\bf X}_{\rm r}^{\rm t}||)$ where ${\bf K}_{\rm p}$ is the proportional gain and ${\bf X}_{\rm G}$ is the goal position.

Figure~\ref{fig: point_robot1} shows the narrow maze where the red dashed line shows the path computed through the maze using the proposed planner. This example is taken from~\cite{chakraborty2009complementarity} where it was solved using a kinodynamic planner. Here, we solved the problem using the proposed kinematic planner. The example demonstrates that the proposed planner can find a feasible path by moving directly towards the goal even when no collision-free direct path exists. This happens as long as there is a non-zero component of the velocity that is tangential to the obstacle surface. The path was computed in $<1$s. 
For comparison, we also solved the problem with modified RRT planner with $10\%$ bias to move towards the goal. We found RRT could not find a path after $10000$ iterations in 6 trials out of 20 trials. The average time RRT took for the $14$ trials to find a path is $66$s with a variance of $15$s. Note that we are not claiming here that for any start-goal pair, we can always reach the goal by directly moving towards it. 
In fact by changing the goal location, our algorithm can get stuck. However, we always get a collision free path and we show later examples, where we use this local planner within a RRT framework.

{\bf Planar 8-DoF  robot:}
Here we present an example using a $8\rm R$ planar robot where the $\mathbb{T}$-space dimension is $2$, the same as our previous example, but the $\mathbb{C}$-space dimension is $8$, which is much higher. The path planning environment is shown in Figure~\ref{fig:snake_plan} with black lines and circles representing the obstacles and robot links as blue lines. The red and green circles are the start and goal position and smaller red circles are robot joints. Our local planner can again find a path by directly moving towards the goal in $2D$ $\mathbb{T}$-space. 
For this example, the time taken to compute the plan is $\sim4$s. The RRT planner could not find a path for this example with corresponding initial and goal joint configurations. However for the same goal position, if the goal joint configuration is altered, RRT found path in average time of $9$s with a standard deviation of $\sim7$s with $20$ trials.
Since mapping of $\mathbb{C}$-space to $\mathbb{T}$-space of redundant manipulators is many to one, for the same goal pose in $\mathbb{T}$-space some joint configurations could be closer to the start joint configuration hence easier to find a path.

{\bf Transferring and pouring liquid while  avoiding obstacle:}
\label{sec:transfer_pour}
\begin{figure*}[!htbp]
    \includegraphics[width=\textwidth]{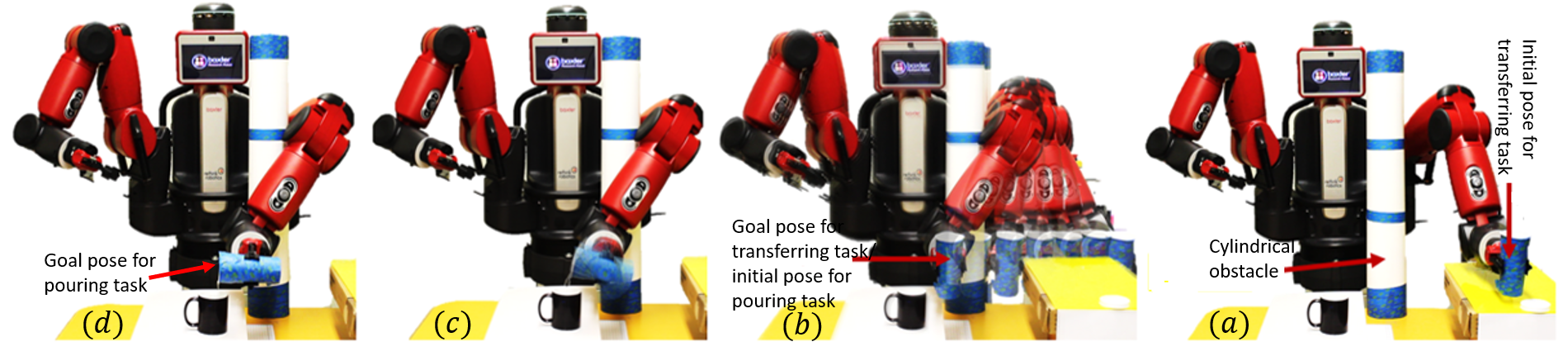}
     \caption{Liquid transferring task as two sub-tasks: {\em Subtask-1} transfer the cup of liquid while avoiding obstacle and maintaining fixed orientation of the end-effector. {\em Subtask-2} pouring the liquid into the mug while keeping position of the end-effector fixed.}
\label{transfer_pour_exp}
\end{figure*}
\begin{figure}
    \centering
    \includegraphics[width=0.5\textwidth]{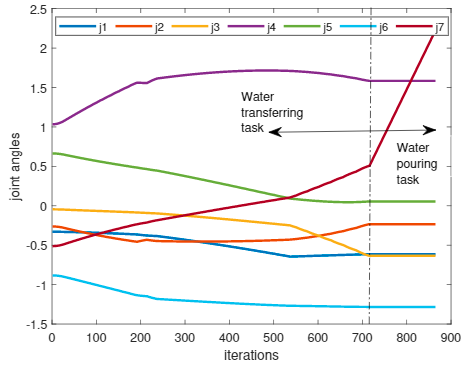}
    \caption{The joint space path corresponding to the task space planning for the water transfer and pouring task as in figure~\ref{transfer_pour_exp}. The legends $j\{i\}$ represent the path of the $i^{th}$ joint $i \in \{1,\dots,7\}$.}
    \label{fig:transfer_pour_exp_plots}
\end{figure}
In this example a plan is computed using the proposed planner for liquid transferring task (see Figure~\ref{transfer_pour_exp}). This example serves to illustrate the following two facts: (a) using ScLERP for interpolating in $\mathbb{T}$-space implicitly satisfy the end-effector constraints for water transferring and pouring tasks. (b) Projection of additional joint-rates $\dot{\bm{\Theta}}_{\rm add}$ required to avoid obstacle into the $Null({\bf J}_{\rm s})$ as presented in Equation~\eqref{eq:null_continuous_model} is helpful to avoid obstacle and maintaining the task constraint simultaneously. To compute the plan, the problem is divided into two sub-problems, one for finding a path for transferring the water and the next to find a path for pouring. For these two planning problem, end-effector's motion is constrained with fixed orientation and fixed position respectively. Using state evolution model as in Equation~\eqref{eq:null_continuous_model} with complementarity based obstacle avoidance paths for the two sub-problems are computed independently and merged at the end to get the full path. The plan for the first sub-task was obtained using the local motion planner with RRT in $\mathbb{T}$-space in $84$s with $715$ nodes added to the tree. The second sub-task was accomplished only using the local planner. In Figure~\ref{transfer_pour_exp}, we show key poses on the computed path. The values of the initial pose, goal pose and initial joint configuration used for this example are as follows. 
For the first sub-task (water transferring task), the orientation of  the start and goal poses of the end-effector is kept fixed while only position is different as shown in ~\ref{transfer_pour_exp}(a) and (b). For the second sub-task (water pouring task) the start and goal end-effector poses has the fixed positio nbut different orientatio nas can be seen in Figure~\ref{transfer_pour_exp}(c) and (d). The fixed orientation constraint and fixed position constraint during the first and second sub-tasks were implicitly handled by ScLERP based interpolation method employed by the proposed planner (see proposition-1 in~\cite{sarker2020}). The joint space path for the computed plan is also presented in Figure~\ref{fig:transfer_pour_exp_plots}. Notice that the path of the joint $7$ changed very little during the water transferring sub-task and the change was large during the pouring sub-task.

\section{Conclusions}
\label{sec:conc}
We have presented a novel kinematic $\mathbb{T}$-space based local planner that uses complementarity constraint to model contact. The proposed planner can be integrated within any sampling based planning framework adapted to task space. The main advantage in modeling contacts using complementarity constraint lies in obtaining a non-zero compensating velocity along the surface normal at any contact point which is then mapped to get a corrected joint rate ensuring all links are collision free. The resultant tangential velocity also helps the robot move along the obstacle surface, thus guiding the robot motion. The usefulness of the proposed planning scheme is evaluated using simulation and experimental results. 
Future work includes an optimized C++ implementation of the planner and integrating it with the vast array of global sampling algorithms proposed in the literature. 
Further, we would like to study $\mathbb{T}$-space based randomized planing scheme with the proposed planer as the local planner. 
\bibliographystyle{asmems4}
\bibliography{ref}
\end{document}